\documentclass[runningheads]{llncs}

\usepackage{eccv}

\usepackage{eccvabbrv}

\usepackage{graphicx}
\usepackage{booktabs}
\usepackage{multirow}
\usepackage{multicol}
\usepackage[accsupp]{axessibility}  

\usepackage{hyperref}

\usepackage{orcidlink}

\begin{document}

\title{Exploring Strengths and Weaknesses of Super-Resolution Attack in Deepfake Detection}

\titlerunning{Super-Resolution Attack in Deepfake Detection}

\author{Davide Alessandro Coccomini\inst{1}\orcidlink{0000-0002-0755-6154} \and
Roberto Caldelli\inst{2,3}\orcidlink{0000-0003-3471-1196} \and
Fabrizio Falchi\inst{1}\orcidlink{0000-0001-6258-5313} \and
Claudio Gennaro\inst{1}\orcidlink{0000-0002-3715-149X} \and
Giuseppe Amato\inst{1}\orcidlink{0000-0003-0171-4315}}

\authorrunning{D. A. Coccomini et al}

\institute{ISTI-CNR, Pisa, Italy \\
\and
CNIT, Florence, Italy \and Mercatorum University, Rome, Italy \\
}

\maketitle

\begin{abstract}
Image manipulation is rapidly evolving, allowing the creation of credible content that can be used to bend reality. Although the results of deepfake detectors are promising, deepfakes can be made even more complicated to detect through adversarial attacks. They aim to further manipulate the image to camouflage deepfakes' artifacts or to insert signals making the image appear pristine.
In this paper, we further explore the potential of super-resolution attacks based on different super-resolution techniques and with different scales that can impact the performance of deepfake detectors with more or less intensity. We also evaluated the impact of the attack on more diverse datasets discovering that the super-resolution process is effective in hiding the artifacts introduced by deepfake generation models but fails in hiding the traces contained in fully synthetic images. Finally, we propose some changes to the detectors' training process to improve their robustness to this kind of attack.
\end{abstract}

\begin{figure*}[t]
\setlength{\abovecaptionskip}{0pt}
  \centering
    \includegraphics[width=\textwidth]{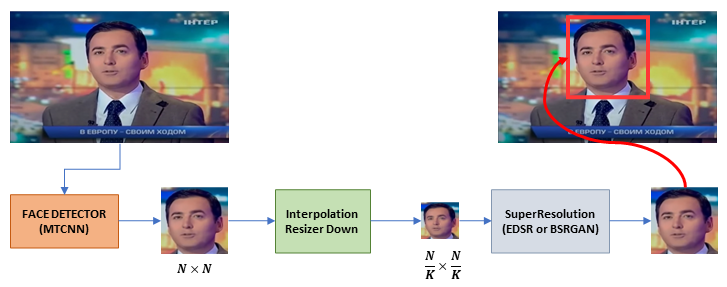}
    \caption{SR attack pipeline: the size of the detected face is reduced by a factor $K$, then restored to its initial resolution using an SR algorithm and pasted back onto the source frame.}
  \label{fig:method}
\end{figure*}

\section{Introduction}

In recent years, the rapid evolution of image generation and manipulation techniques has enabled individuals to create extremely convincing visual content, creating an environment in which the line between reality and fiction becomes increasingly thin. Parallel to this development, numerous deepfake detection tools have been developed, using diverse approaches to mitigate the phenomenon and accurately distinguish between authentic and manipulated images. However, despite promising advances in such detectors, the persistence of deepfakes remains a challenge, especially considering the use of advanced adversarial attack techniques.
Adversarial attacks aim to further complicate the detection process by manipulating the image in ways that are often imperceptible to the human eye with the goal of hiding the traces of deepfakes or inserting specific signals that can mislead the deepfake detectors into believing that the image (or video) is authentic.

\noindent In previous research \cite{coccominiadversarial}, it has been shown that the use of super-resolution (SR) techniques effectively acts as a test-time adversarial attack, blurring the artifacts introduced by deepfakes and making them extremely difficult to identify. At the same time, the usage of super-resolution on pristine images can make them be confused as fake from deepfake detectors, causing a huge amount of false alarms. However, the impact of super-resolution on deepfake detectors' performances has been little explored.
This paper aims to further investigate the potential of super-resolution-based attacks, examining whether this attack is always effective regardless of the type of super-resolution technique used and on different kinds of manipulated images. Also, we explore the impact of different super-resolution configurations both in terms of the effectiveness of the attack and in terms of impact on the image's visual appearance. Through this in-depth analysis, we aim to provide a more comprehensive understanding of how such attacks can make the task of detecting manipulated images even more challenging. 
We were able to see how using super-resolution techniques can always distort images enough to damage the performance of deepfake detectors, independently from the kind of super-resolution approach used. However, the attack is ineffective when applied to fully synthetically generated images. We, therefore, believe that the action of super-resolution is to camouflage artifacts introduced by manipulations made on real images but is not capable of making entirely synthetic images appear as pristine. 
Furthermore, we have verified that in order to make the attack more effective, it is possible to increase the scale at which super-resolution is performed and thus manipulate the image resolution to a greater extent, although this has potentially considerable costs in terms of the visual appearance of the resulting image. In this work, we also investigated some additional training procedures to improve deepfake detectors' robustness. Summarizing, we try to answer the following research questions:

\begin{itemize}
    \item is the SR-based attack effective on deepfake detectors regardless of the type of SR method used?
    \item is SR-based attack also effective on entirely synthetic images?
    \item how does the scale of super-resolution application affect both the visual quality of the resulting image and the performance of deepfake detectors?
    \item how can we improve the training process in order to make the deepfake detector more robust to the SR-based attack?
\end{itemize}

\section{Related Works}
\subsection{Deepfake Detection and Adversarial attacks}
As deepfakes gain credibility, detecting them becomes crucial and this has led to the development of several image and video deepfake detectors. Some of them can be applied to videos considering both spatial and temporal information \cite{mintime,Baxevanakis2022TheMD,CALDELLI202131} while some other frame-based methods \cite{coccominicombining} can be used also on images. The increase of interest in this field conducted also to the launch of several competitions like \cite{dolhansky2020deepfake} and \cite{jimaging8100263} further stimulating advancements by challenging researchers to find innovative ways of detecting deepfakes. Detection efforts have also recently extended to fully synthetic images with research works like \cite{coccominidiffusion,Dogoulis_2023,amoroso2023parents} which demonstrated how it is feasible to detect synthetic images with similar problems that can be found in traditional deepfake detection.\\
Over the years, many approaches have also been developed to attack deepfake detectors and make the detection even more challenging. For example, approaches like noise addition and adversarial patches introduce subtle perturbations or overlap patterns to trigger misclassification. The method called FakeRetouch \cite{huang2020fakeretouch} reduces deepfake artifacts without compromising image quality. They attain a remarkable fidelity to the initial deepfake images by incorporating noise and employing deep image filtering, thereby diminishing the accuracy of deepfake detectors. A pretty different approach is StatAttack \cite{Hou_2023_CVPR} which minimizes statistical differences between deepfake images and natural ones to deceive detectors by means of the addition of statistical-sensitive degradations.

\subsection{Super-Resolution}
Super-resolution (SR) reconstructs high-resolution images from low-resolution ones using multiple input images \cite{arefin2020multi} or prior knowledge \cite{8723565}. A successful example of super-resolution approach is EDSR \cite{Lim_2017_CVPR_Workshops}: it enhances the ResNet architecture removing batch normalization for flexibility and introducing residual scaling layers for stability in order to create a model capable of increasing the resolution of an image. 
Another relevant proposal has been made in \cite{bsrgan} which proposes a powerful practical degradation model which uses shuffled blur, downsampling and noise degradations, to effectively train several models on pristine-degraded images to perform the super-resolution task, in particular BSRGAN.

\begin{table*}[t]
    \centering
    \setlength{\tabcolsep}{0.7em}
    \resizebox{\textwidth}{!}{%
    \begin{tabular}{l|l|r|r|r|r|r|r|r|r}
    \hline\hline
    \textbf{Model} & \textbf{Forgery Method} & \textbf{SR} & \textbf{SR Method} & \textbf{FNR} $(\%)$$\downarrow$ & \textbf{FPR} $(\%)$$\downarrow$ & \textbf{Recall} $(\%)$$\uparrow$& \textbf{Precision} $(\%)$$\uparrow$ & \textbf{AUC}$(\%)$ $\uparrow$ & \textbf{Accuracy} $(\%)$$\uparrow$ \\
\hline
\multirow{15}{*}{\textbf{Resnet50}} & \multirow{3}{*}{Deepfakes} & $\times$ & None & 5.5 & 3.2 & 94.5 & 96.7 & 99.2 & 95.6 \\
& & $\checkmark$ & EDSR & 6.9 & 10.1 & 93.1 & 90.2 & 97.7 & 91.5 \\
&  & $\checkmark$ & BSRGAN & 15.5 & 18.6 & 84.5 & 82.0 & 91.9 & 83.0\\
\cline{2-10}
& \multirow{3}{*}{Face2Face} & $\times$ & None & 5.0 & 3.2 & 95.0 & 96.7 & 98.9 & 95.9 \\
& & $\checkmark$ & EDSR & 14.4 & 4.7 & 85.6 & 94.8 & 97.0 & 90.5 \\
&  & $\checkmark$ & BSRGAN & 27.4 & 10.2 & 72.6 & 87.7 & 90.8 & 81.2\\
\cline{2-10}
& \multirow{3}{*}{FaceSwap}  & $\times$ & None & 6.4 & 1.9 & 93.6 & 98.0 & 99.1 & 95.9 \\
& & $\checkmark$ & EDSR & 21.1 & 2.6 & 78.9 & 96.8 & 96.0 & 88.1 \\
&  & $\checkmark$ & BSRGAN & 39.9 & 4.9 & 60.1 & 92.4 & 89.2 & 77.6\\
\cline{2-10}
& \multirow{3}{*}{FaceShifter} & $\times$ & None & 6.1 & 3.4 & 93.9 & 96.5 & 98.8 & 95.3 \\
& & $\checkmark$ & EDSR & 24.8 & 3.3 & 75.2 & 95.8 & 96.8 & 86.0 \\
&  & $\checkmark$ & BSRGAN & 66.6 & 9.0 & 33.4 & 78.8 & 84.6 & 62.2\\
\cline{2-10}
& \multirow{3}{*}{NeuralTextures} & $\times$ & None & 14.1 & 8.1 & 85.9 & 91.3 & 95.4 & 88.9 \\
& & $\checkmark$ & EDSR & 14.4 & 16.9 & 85.6 & 83.5 & 92.1 & 84.4 \\
&  & $\checkmark$ & BSRGAN & 32.9 & 28.4 & 67.1 & 70.3 & 77.6 & 69.4\\
\hline
\multirow{15}{*}{\textbf{Swin}} & \multirow{3}{*}{Deepfakes} & $\times$ & None & 5.9 & 3.6 & 94.1 & 96.3 & 99.1 & 95.3 \\
& & $\checkmark$ & EDSR & 6.1 & 12.4 & 93.9 & 88.4 & 97.4 & 90.7 \\
&  & $\checkmark$ & BSRGAN & 29.5 & 9.3 & 70.5 & 88.4 & 90.9 & 80.6\\
\cline{2-10}
& \multirow{3}{*}{Face2Face} & $\times$ & None & 6.3 & 3.3 & 93.7 & 96.6 & 98.9 & 95.2 \\
&  & $\checkmark$ & EDSR & 24.4 & 1.7 & 75.6 & 97.8 & 96.1 & 87.0 \\
&  & $\checkmark$ & BSRGAN & 34.1 & 5.2 & 65.9 & 92.7 & 91.5 & 80.4\\
\cline{2-10}
& \multirow{3}{*}{FaceSwap} & $\times$ & None & 4.9 & 4.6 & 95.1 & 95.3 & 98.6 & 95.2 \\
&  & $\checkmark$ & EDSR & 21.9 & 5.3 & 78.1 & 93.7 & 93.9 & 86.4 \\
&  & $\checkmark$ & BSRGAN & 45.9 & 5.8 & 54.1 & 90.3 & 87.1 & 74.2\\
\cline{2-10}
& \multirow{3}{*}{FaceShifter} & $\times$ & None & 7.2 & 4.1 & 92.8 & 95.8 & 98.7 & 94.4 \\
&  & $\checkmark$ & EDSR & 18.9 & 3.1 & 81.1 & 96.3 & 97.4 & 89.0 \\
&  & $\checkmark$ & BSRGAN & 52.1 & 11.8 & 47.9 & 80.3 & 84.7 & 68.1\\
\cline{2-10}
& \multirow{3}{*}{NeuralTextures} & $\times$ & None & 12.9 & 12.8 & 87.1 & 87.2 & 94.9 & 87.1 \\
&  & $\checkmark$ & EDSR & 13.2 & 23.9 & 86.8 & 78.4 & 90.4 & 81.5 \\
&  & $\checkmark$ & BSRGAN & 43.0 & 24.2 & 57.0 & 70.2 & 74.0 & 66.4\\
\hline
\multirow{15}{*}{\textbf{XceptionNet}} & \multirow{3}{*}{Deepfakes} & $\times$ & None & 5.3 & 2.6 & 94.7 & 97.4 & 99.3 & 96.1 \\
& & $\checkmark$ & EDSR & 5.6 & 12.4 & 94.4 & 88.4 & 97.9 & 91.0 \\
&  & $\checkmark$ & BSRGAN & 14.2 & 14.4 & 85.8 & 85.6 & 94.0 & 85.7\\
\cline{2-10}
& \multirow{3}{*}{Face2Face} & $\times$ & None & 9.6 & 3.3 & 90.4 & 96.5 & 98.4 & 93.6 \\
& & $\checkmark$ & EDSR & 18.3 & 5.3 & 81.7 & 93.9 & 95.7 & 88.2 \\
&  & $\checkmark$ & BSRGAN & 22.4 & 19.8 & 77.6 & 79.7 & 88.4 & 78.9\\
\cline{2-10}
& \multirow{3}{*}{FaceSwap} & $\times$ & None & 5.1 & 3.2 & 94.9 & 96.7 & 98.8 & 95.8 \\
&  & $\checkmark$ & EDSR & 15.8 & 4.9 & 84.2 & 94.5 & 96.6 & 89.6 \\
&  & $\checkmark$ & BSRGAN & 23.9 & 8.9 & 76.1 & 89.5 & 92.1 & 83.6\\
\cline{2-10}
& \multirow{3}{*}{FaceShifter} & $\times$ & None & 7.1 & 3.9 & 92.9 & 96.0 & 98.8 & 94.5 \\
&  & $\checkmark$ & EDSR & 15.6 & 4.3 & 84.4 & 95.2 & 97.4 & 90.0 \\
&  & $\checkmark$ & BSRGAN & 56.4 & 8.5 & 43.6 & 83.7 & 86.3 & 67.5\\
\cline{2-10}
& \multirow{3}{*}{NeuralTextures} & $\times$ & None & 13.1 & 7.2 & 86.9 & 92.3 & 95.9 & 89.8 \\
& & $\checkmark$ & EDSR & 9.8 & 21.6 & 90.2 & 80.7 & 92.7 & 84.3 \\
&  & $\checkmark$ & BSRGAN & 40.1 & 23.4 & 59.9 & 72.0 & 76.2 & 68.3\\\hline
\hline
    
    \end{tabular}
    }
     \caption{Evaluation on FF++ test set (half pristine and half fake). The SR column indicates if the SR adversarial technique has been applied to the images. \textbf{Both pristine and fake images are attacked with SR using $K=2$.}}
\label{tab:sr_comparison}
\end{table*}

\section{The Proposed Attack Procedure}
\label{sec:proposed_method}
To perform the super-resolution attack we followed the strategy presented in \cite{coccominiadversarial} and the proposed implementation framework is illustrated in Figure \ref{fig:method}. The attack is designed to apply the super-resolution to reduce the possible artifacts introduced by the deepfake generation process and eventually learnt by a trained deepfake detector, thus making detection more complex.
To remain consistent with the previous work done in this field \cite{coccominiadversarial}, for the main experiments we use a video deepfake dataset and for each frame in a video slated for analysis, a pretrained face detector (e.g., MTCNN \cite{mtcnn}) is initially adopted. This addition to the pipeline serves two crucial purposes. Firstly, it aligns with the attack's objective, as an attacker aims to manipulate only a minimal portion of the image to avoid introducing unnecessary artifacts. Applying SR to the entire frame might inadvertently introduce artifacts to the background, resulting in counterproductive effects. Secondly, the utilization of a face detector is coherent with the common practice of both deepfake detectors and generators, which typically focus on facial features. Consequently, the targeted deepfake detector will probably concentrate on facial aspects and artifacts to be removed will likely be concentrated on the face too.
The face extracted by the face detector can have a specific resolution determined by factors such as video resolution and the person's distance from the camera. Given that SR aims to enhance image resolution by a factor $K \in \mathbb{N}$, the image is initially downscaled by a factor of $1/K$ and then inputted into an SR model (e.g., EDSR \cite{Lim_2017_CVPR_Workshops}) for upscaling by a factor of $K$ (see Figure \ref{fig:method}). The resulting face image from this process maintains the same size as the originally detected one and can be seamlessly reintegrated into the source image from which it had been taken.
Importantly, this method does not necessitate any knowledge about the specific deepfake detector to be employed for final detection. Thus, the proposed approach can be effectively considered a black-box attack, applicable across various deepfake detectors and on images manipulated using any deepfake generation method. Furthermore, this attack can be executed on pre-existing deepfake content, eliminating the need for integration into the deepfake creation procedure.

\noindent Differently to what has been done in \cite{coccominiadversarial}, we explore diverse super-resolution techniques, applied to more varied data and with a set of different scale factors, in order to more widely validate the effectiveness of SR-based attacks. We also propose some additional training procedures to improve the detector's robustness.

\section{Experiments}
\subsection{Experimental Setup}

To assess the influence of super-resolution (SR) application on the performance of deepfake detectors, we opted for three architectures: Resnet50, Swin-Small, and XceptionNet. These models were trained on faces extracted from the well-known FaceForensics++ (FF++) dataset \cite{rossler2019faceforensics} in the c23 version, for binary classification of pristine/fake images. During training, each model exclusively encountered pristine and manipulated fake images using one of the five available forgery methods (e.g. see the second column of Table \ref{tab:sr_comparison} for such methods) without applying super-resolution. All models were pre-trained on ImageNet, followed by fine-tuning with a learning rate of 0.01 for 30 epochs on an Nvidia Tesla T4. During fine-tuning, some basic data augmentations techniques are randomly applied such as noise addition, image compression, geometric transformations etc.

\noindent In general, models were tested with the SR-attack applied to both fake and pristine images. A pretrained MTCNN \cite{mtcnn} was employed to extract faces from each frame. Our experiments utilized a variable scale factor, resizing the extracted face by a factor of $1/K$ before upscaling through EDSR \cite{Lim_2017_CVPR_Workshops} or BSRGAN \cite{bsrgan} to restore the initial resolution as previously explained in Section \ref{sec:proposed_method}.

\noindent We then conducted the same experiments but using face images entirely generated with various synthetic image generation methods and applying the attack on fake images only. In particular, we used a dataset composed of images generated with StyleGAN, StyleGAN2, RelGAN and ProGAN, presented in \cite{uninagans}. We tested the pretrained models provided by the authors which were trained on a dataset composed by pristine and StyleGAN2 images created by the authors or ProGAN images taken from \cite{progandataset} and we validated their robustness to the SR-based attack.

\begin{table}[t]
\centering
\vspace{1mm}
\setlength{\tabcolsep}{4pt}
\resizebox{0.7\textwidth}{!}{%
\begin{tabular}{l|c|c|c|c}
\hline
\hline
\textbf{Forgery Method} & \textbf{Scale ($K$)} & \textbf{SR Method} & \textbf{SSIM} $\uparrow$ & \textbf{PSNR $\uparrow$ (dB)} \\
\hline\hline
\multirow{4}{*}{\emph{Pristine}} & x2 & \multirow{2}{*}{EDSR} & 0.968 & 39.8 \\
& x4 & & 0.910 & 34.8 \\
\cline{2-5}
& x2 & \multirow{2}{*}{BSRGAN} & 0.937 & 35.7 \\
& x4 & & 0.858 & 32.9 \\
\hline
\multirow{4}{*}{Deepfakes} & x2 & \multirow{2}{*}{EDSR} & 0.970 & 40.3 \\
& x4 & & 0.917 & 34.1 \\
\cline{2-5}
& x2 & \multirow{2}{*}{BSRGAN} & 0.938 & 35.8 \\
& x4 & & 0.860 & 33.0  \\
\hline
\multirow{4}{*}{Face2Face} & x2 & \multirow{2}{*}{EDSR} & 0.968 & 39.8 \\
& x4 & & 0.912 & 33.9 \\
\cline{2-5}
& x2 & \multirow{2}{*}{BSRGAN} & 0.938 & 35.9 \\
& x4 & & 0.859 & 33.0 \\
\hline
\multirow{4}{*}{FaceShifter} & x2 & \multirow{2}{*}{EDSR} & 0.973 & 40.9 \\
& x4 & & 0.927 & 34.4 \\
\cline{2-5}
& x2 & \multirow{2}{*}{BSRGAN} & 0.936 & 35.6 \\
& x4 & & 0.866 & 33.1 \\
\hline
\multirow{4}{*}{FaceSwap} & x2 & \multirow{2}{*}{EDSR} & 0.967 & 39.7\\
& x4 & & 0.910 & 33.9 \\
\cline{2-5}
& x2 & \multirow{2}{*}{BSRGAN} & 0.938 & 35.9\\
& x4 & & 0.859 & 33.0 \\
\hline
\multirow{4}{*}{NeuralTextures} & x2 & \multirow{2}{*}{EDSR} & 0.972 & 40.5 \\
& x4 & & 0.921 & 34.2 \\
\cline{2-5}
& x2 & \multirow{2}{*}{BSRGAN} & 0.937 & 35.7 \\
& x4 & & 0.866 & 33.0 \\
\hline
\hline
\end{tabular}
}
\caption{Evaluation of perceptual similarity (SSIM and PSNR) between non-SR images and SR ones for each forgery method and for the pristine case.}
\label{tab:similarity}
\end{table}

\begin{table}[t]
\centering

\resizebox{\textwidth}{!}{%
\begin{tabular}{c|c|c|c|c|c|c}
\hline
\hline
\multirow{2}{*}{\textbf{Forgery Method}} & \multirow{2}{*}{\textbf{SR Method}} & \multirow{2}{*}{\textbf{Scale ($K$)}} & \multirow{2}{*}{\textbf{FNR $(\%)$$\downarrow$ }}  & \multirow{2}{*}{\textbf{FPR $(\%)$$\downarrow$}} & \multirow{2}{*}{\textbf{AUC $(\%)$$\uparrow$}}  & \multirow{2}{*}{\textbf{Accuracy $(\%)$$\uparrow$}}  \\
& & & & & & \\
\hline
\hline

\multirow{2}{*}{Deepfakes} & \multirow{10}{*}{EDSR} & x2 & 6.9 & 10.1  & 97.7 & 91.5 \\
& & x4 & 5.9 & 41.3 & 91.4 & 76.4 \\
\cline{1-1}
\cline{3-7}
\multirow{2}{*}{Face2Face} & & x2 & 14.4 & 4.7 & 97.0 & 90.5 \\
 & & x4 & 49.9 & 5.8 & 85.8 & 72.2 \\
\cline{1-1}
\cline{3-7}
\multirow{2}{*}{FaceSwap} & & x2 & 21.1 & 2.6 & 96.0 & 88.1 \\
& & x4 & 68.6 & 2.9 & 82.7 & 64.2 \\
\cline{1-1}
\cline{3-7}
\multirow{2}{*}{FaceShifter} & & x2 & 24.8 & 3.3 & 96.8 & 86.0 \\
&  & x4 & 16.5 & 7.9 & 95.6 & 87.8 \\
\cline{1-1}
\cline{3-7}
\multirow{2}{*}{NeuralTextures} & & x2 & 14.4 & 16.9 & 92.1 & 84.4 \\
& & x4 & 4.9 & 61.4 & 84.1 & 66.9 \\
\hline\hline
\multirow{2}{*}{Deepfakes} & \multirow{10}{*}{BSRGAN} & x2 & 15.5 & 18.6 & 91.9 & 83.0 \\
& & x4 & 39.7 & 16.3 & 82.5 & 72.0 \\
\cline{1-1}
\cline{3-7}
\multirow{2}{*}{Face2Face} & & x2 & 27.4 & 10.2 & 90.8 & 81.2 \\
& & x4 & 41.9 & 16.0 & 79.7 & 71.0 \\
\cline{1-1}
\cline{3-7}
\multirow{2}{*}{FaceSwap} & & x2 & 39.9 & 4.9 & 89.2 & 77.6 \\
& & x4 & 66.1 & 6.0 & 72.0 & 64.0 \\
\cline{1-1}
\cline{3-7}
\multirow{2}{*}{FaceShifter} & & x2 & 66.6 & 9.0 & 84.6 & 62.2 \\
& & x4 & 53.8 & 31.4 & 64.7 & 57.4 \\
\cline{1-1}
\cline{3-7}
\multirow{2}{*}{NeuralTextures} & & x2 & 32.9 & 28.4 & 77.6 & 69.4 \\
& & x4 & 53.8 & 26.3 & 65.7 & 60.0 \\
\hline
\hline

\end{tabular}

}
\caption{The impact of scale factor $K$ in the SR-based attack on Resnet50-based deepfake detector's performances. The same behaviour is obtained for Swin-Base and XceptionNet. Evaluated on FF++ test set. \textbf{Both pristine and fake images are attacked with SR.}}
\label{tab:scale_performance_impact}
\end{table}

\begin{table}[t]
    \centering
    \resizebox{0.6\textwidth}{!}{%
    \begin{tabular}{c|c|c|c}
    \hline
    \hline
         \textbf{Test Set}  & \textbf{Training} Set & \textbf{Scale ($K$)} & \textbf{Accuracy $(\%)$$\uparrow$}\\
         \hline\hline
 &  & None & 99.9\\
ProGAN (CelebAhq) & \multirow{12}{*}{StyleGAN2} & x2 & 99.9\\
 &  & x4 & 98.5\\
\cline{1-1}\cline{3-4}
&  & None & 99.6\\
RelGAN (CelebA) & & x2 & 99.9\\
&  & x4 & 98.8\\
\cline{1-1}\cline{3-4}
&  & None & 100.0\\
StyleGAN (FFHQ) & & x2 & 100.0\\
&  & x4 & 100.0\\
\cline{1-1}\cline{3-4}
&  & None & 100.0\\
StyleGAN2 (FFHQ) & & x2 & 100.0\\
&  & x4 & 100.0\\
\cline{1-1}\cline{3-4}
&  & None & 100.0\\
StyleGAN (CelebAhq) & & x2 & 100.0\\
&  & x4 & 98.7\\
\hline\hline
&  & None & 100.0\\
ProGAN (CelebAhq) & \multirow{12}{*}{ProGAN} & x2 & 100.0\\
&  & x4 & 100.0\\
\cline{1-1}\cline{3-4}
&  & None & 96.1\\
RelGAN (CelebA) & & x2 & 99.5\\
&  & x4 & 97.5\\
\cline{1-1}\cline{3-4}
&  & None & 99.1\\
StyleGAN (FFHQ) & & x2 & 97.3\\
&  & x4 & 96.7\\
\cline{1-1}\cline{3-4}
&  & None & 99.6\\
StyleGAN2 (FFHQ) & & x2 & 99.5\\
&  & x4 & 99.6\\
\cline{1-1}\cline{3-4}
&  & None & 100.0\\
StyleGAN (CelebAhq) & & x2 & 100.0\\
&  & x4 & 100.0\\
\hline
 
\hline\hline
    \end{tabular}
    }
    \caption{Evaluation of a Resnet50-based detector on Syntethic images dataset \cite{uninagans}. Swin-Base and XceptionNet behave similarly. The test set column indicates the kind of GAN network and its training set used to create the test set. \textbf{Only the fake images are attacked with SR using EDSR.}}
    \label{tab:evaluation_gan}
\end{table}

\subsection{Evaluation of SR-based attack on deepfake images}
In \cite{coccominiadversarial}, it was verified that the use of an SR-based attack which exploits EDSR is effective in camouflaging deepfakes artifacts obtained by various methodologies. In this section, we test the effectiveness of the attack using a different technique, namely BSRGAN \cite{bsrgan}.
In Table \ref{tab:sr_comparison}, we compare the impact of EDSR- or BSRGAN-based attacks on the performance of three different deepfake detectors on the FaceForensics++ dataset. As the table shows, the effectiveness of the attack is confirmed also using BSRGAN, determining a significant increment in percentage of False Negative Rate (FNR) and False Positive Rate (FPR) regardless of the super-resolution method used; moreover, it appears that a more remarkable impact is registered when the BSRGAN is used as the SR method. On the \emph{Deepfakes} method, the attack has a greater effect on the FPR by making pristine images appear as fakes in the eyes of the detector. On \emph{Face2Face}, \emph{FaceSwap} and \emph{FaceShifter} we obtain the opposite tendency and the attack works very well in making deepfake images difficult to recognize, thus drastically raising the FNR. Finally, on \emph{NeuralTextures} there is an increase in both the FNR and the FPR in almost all the considered contexts. The other performance measures such as Precision, Recall, Area-Under-Curve (AUC) and Accuracy consequently highlight the same general tendency.   

\begin{figure*}[t]
\setlength{\abovecaptionskip}{0pt}
  \centering
    \includegraphics[width=0.8\textwidth]{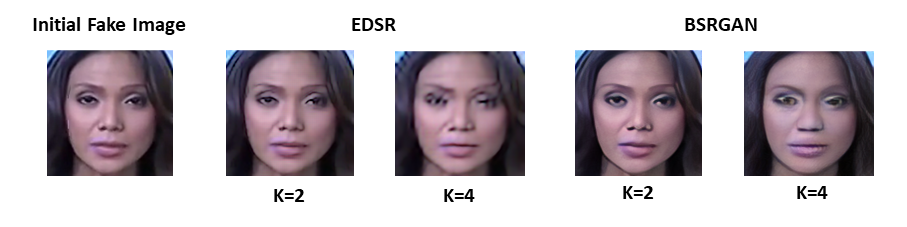}
    \caption{Example of a fake image before and after the SR-attack applied with different SR methods and scales.}
  \label{fig:sr_samples}
\end{figure*}

\subsection{Impact of scale factor}
From the previous experiments, BSRGAN seems to be more effective than EDSR in causing more pronounced decrement on detector performance, however, looking at Table \ref{tab:similarity} in which we report the perceptual similarity (SSIM and PSNR) between dataset's images before and after the SR attack, the changes made by BSRGAN-based attack seems to be slightly more visible and this is evidenced by the similarity scores with respect to the non-SR images. Also, as expected, processing images through super-resolution using a higher scale factor always leads to a visual result that is less similar to the source image and therefore may make the manipulation more identifiable to the naked eye while it could be more effective in covering the artifacts learned by the detector. In Figure \ref{fig:sr_samples} it can be seen that using an SR-attack based on EDSR or BSRGAN with a scale factor $K=2$, removes the typical artifacts introduced by deepfake generators (such as in the example, the net line on the cheek of the girl). The usage of a scale factor $K=4$ instead excessively distorts the image, resulting in both the removal of the artifacts introduced by the deepfake generators and the introduction of other artifacts that make it visually strange. For that reason, it is generally suggested to apply the attack with a low scale factor otherwise the distortion may be excessive.

\noindent In that sense, in Table \ref{tab:scale_performance_impact} we report the performance impact of a Resnet50-based deepfake detector when using a larger or smaller scale factor. As it can be seen from the results, as the parameter $K$ increases, the performance of the deepfake detector deteriorates, causing the FPR and/or FNR to increase more and more. This is perfectly in line with previous experiments as this performance decay is related to how much the attack modifies the image and the higher the $K$ parameter the more the image will be altered.

\begin{table*}[t]
    \centering
    \setlength{\tabcolsep}{0.7em}
    \resizebox{\textwidth}{!}{
    \begin{tabular}{l|l||r|r||r|r||r|r||r|r}
    \hline\hline
    \multirow{2}{*}{\textbf{Forgery Method}} & \multirow{2}{*}{\textbf{SR Method}} & \multicolumn{2}{c||}{\textbf{FNR $(\%)$$\downarrow$}} & \multicolumn{2}{|c||}{\textbf{FPR $(\%)$$\downarrow$}} & \multicolumn{2}{|c||}{\textbf{AUC $(\%)$$\uparrow$}} & \multicolumn{2}{|c}{\textbf{Accuracy $(\%)$$\uparrow$}} \\
    \cline{3-4} \cline{5-6} \cline{7-8} \cline{9-10}
    & & \textbf{w/o SR Aug} & \textbf{w/ SR Aug} & \textbf{w/o SR Aug} & \textbf{w/ SR Aug} & \textbf{w/o SR Aug} & \textbf{w/ SR Aug} & \textbf{w/o SR Aug} & \textbf{w/ SR Aug} \\
    \hline
    \multirow{3}{*}{Deepfakes} & None & \textbf{5.5} & 5.7 & 3.2 & \textbf{2.6} & 99.2 & 99.2 & 95.6 & \textbf{95.9} \\
    & EDSR & \textbf{6.9} & 7.6 & \textbf{10.1} & 10.4 & \textbf{97.7} & 97.6 & \textbf{91.5} & 91.0 \\
    & BSRGAN & \textbf{15.5} & 16.4 & 18.6 & \textbf{14.4} & 91.9 & \textbf{92.9} & 83.0 & \textbf{84.6} \\
    \hline
    \multirow{3}{*}{Face2Face} & None & \textbf{5.0} & 7.7 & \textbf{3.2} & 5.6 & \textbf{98.9} & 98.4 & \textbf{95.9} & 93.3 \\
    & EDSR & 14.4 & \textbf{9.9} & \textbf{4.7} & 7.2 & 97.0 & \textbf{97.3} & 90.5 & \textbf{91.5} \\
    & BSRGAN & 27.4 & \textbf{15.4} & 10.2 & \textbf{9.4} & 90.8 & \textbf{95.2} & 81.2 & \textbf{87.6} \\
    \hline
    \multirow{3}{*}{FaceSwap} & None & 6.4 & \textbf{5.9} & \textbf{1.9} & 3.6 & \textbf{99.1} & 98.8 & \textbf{95.9} & 95.3 \\
    & EDSR & 21.1 & \textbf{8.0} & \textbf{2.6} & 4.3 & 96.0 & \textbf{98.1} & 88.1 & \textbf{93.9} \\
    & BSRGAN & 39.9 & \textbf{10.6} & 4.9 & \textbf{3.9} & 89.2 & \textbf{97.5} & 77.6 & \textbf{92.7} \\
    \hline
    \multirow{3}{*}{FaceShifter} & None & \textbf{6.1} & 6.7 & 3.4 & \textbf{3.2} & 98.8 & \textbf{99.0} & \textbf{95.3} & 95.0 \\
    & EDSR & 24.8 & \textbf{10.2} & 3.3 & \textbf{2.3} & 96.8 & \textbf{98.7} & 86.0 & \textbf{93.8} \\
    & BSRGAN & 66.6 & \textbf{11.2} & 9.0 & \textbf{4.1} & 84.6 & \textbf{98.3} & 62.2 & \textbf{92.4} \\
    \hline
    \multirow{3}{*}{NeuralTextures} & None & 14.1 & \textbf{13.7} & 8.1 & 8.1 & 95.4 & 95.4 & 88.9 & \textbf{89.1} \\
    & EDSR & \textbf{14.4} & 15.4 & 16.9 & \textbf{11.9} & 92.1 & \textbf{93.5} & 84.4 & \textbf{86.4} \\
    & BSRGAN & 32.9 & \textbf{24.9} & 28.4 & \textbf{24.3} & 77.6 & \textbf{83.7} & 69.4 & \textbf{75.4} \\
    \hline\hline
    \end{tabular}
    }
    \caption{Comparison of Resnet50 performance with and without SR-based data augmentation. The results are similar for Swin-Base and XceptionNet. \textbf{Both pristine and fake images are attacked with SR using $K=2$.}}
    \label{tab:comparison_sr_aug}
\end{table*}

\subsection{Evaluation of SR-based attack on synthetic images}
In the previous experiments, we have seen how it is possible to use an SR-based adversarial attack to deteriorate the performance of a deepfake detector on images manipulated with classical deepfake generation techniques (e.g. \emph{FaceSwap}, \emph{Face2Face}, etc.).
In this section, we try to examine the effectiveness of the proposed attack on images which are entirely synthetic, i.e. created using Generative Adversarial Networks (GANs) of various types. 
In Table \ref{tab:evaluation_gan} we report the results obtained by a Resnet50-based detector, trained and tested on a dataset of synthetic images \cite{uninagans}. The results show that the attack is ineffective regardless of the scale at which it is applied. This probably lies in the fact that classic deepfake methods (as in FF++ dataset) tend to introduce specific artifacts that are camouflaged by the super-resolution process, whereas in fully synthetic images the traces of the generation are present more widely in the image and continue to be recognisable by the detector even after the SR attack. As demonstrated in \cite{10209003} in fact, in fully synthetic images it is possible to individuate artifacts and patterns in the frequency domain in addition to simply visual artifacts.

\subsection{Contrast the SR-based attack}
The possibility of having the performance of deepfake detectors deteriorate with the simple use of an SR-based attack highlights how important it is to make detectors robust to this type of distortion. In the following, we therefore propose training a deepfake detector by introducing super-resolution techniques into the data augmentation. 
We have therefore trained a Resnet50 on each deepfake generation method present in FaceForensics++ but each time an image is selected for batch construction, it can be subjected to EDSR or BSRGAN super-resolution, with $K=2$ or $K=4$. 
In Table \ref{tab:comparison_sr_aug} we report the performance comparison between a Resnet50 trained with and without the use of this data augmentation technique on the test set. As can be seen from the results, introducing super-resolution techniques into the data augmentation in the training phase tends to lead to greater robustness to SR-attack, which can be more or less pronounced depending on the type of deepfake generation method used to create the initial fake image. In particular, the accuracy is almost always improved thanks to this data augmentation approach and there are some cases in which we obtained a huge drop in FNR such as FaceShifter. Despite this, the SR-based attack continues to be fairly effective but the use of this specific data augmentation may be a good solution to counter this attack and make the deepfake detectors more robust.

\section{Conclusions}
In this paper, we investigated the use of super-resolution-based attacks against deepfake detectors by examining their effectiveness both on real images manipulated by deepfake generation techniques, where the attack was particularly effective, and on synthetic images generated by GAN where the attack failed. Furthermore, we experimented with attacks based on different types of super-resolution methods and examined the impact of scaling on the performance and visual appearance of the resulting images. 
SR-based adversarial attacks appear to be particularly effective in the field of deepfake detection and it is therefore crucial to design the deepfake detectors of the future by also seeking robustness to this type of processing. For that reason, we also proposed an effective technique to train deepfake detectors to make them more robust to SR-based attacks obtaining promising results. As a future work, we will continue exploring different ways to improve the SR-based attack such as the application of the SR to the whole frame and also further explore how to make deepfake detectors more robust to those situations.

\section{Acknowledgments}
This work was partially supported by the following projects:
Tuscany Health Ecosystem (THE) (CUP B83C22003930001) and 
SERICS (PE00000014) under the MUR National Recovery and Resilience Plan funded by European Union - NextGenerationEU,
AI4Debunk (GA n. 101135757) funded by the EU Horizon Europe Programme,
FOSTERER project funded by the Italian MUR (PRIN 2022).

\bibliographystyle{splncs04}
\bibliography{main}
\end{document}